\documentclass[sigconf]{acmart}
\AtBeginDocument{%
  \providecommand\BibTeX{{%
    \normalfont B\kern-0.5em{\scshape i\kern-0.25em b}\kern-0.8em\TeX}}}

\setcopyright{acmcopyright}
\copyrightyear{2018}
\acmYear{2018}
\acmDOI{XXXXXXX.XXXXXXX}

\acmConference[Conference acronym 'XX]{Make sure to enter the correct
  conference title from your rights confirmation emai}{June 03--05,
  2018}{Woodstock, NY}
\acmPrice{15.00}
\acmISBN{978-1-4503-XXXX-X/18/06}

\usepackage{multirow}
\usepackage{hhline}  
\usepackage{comment}

\usepackage{listings} 
\usepackage{xcolor}
\usepackage{CJKutf8}

\definecolor{codegreen}{rgb}{0,0.6,0}
\definecolor{codegray}{rgb}{0.5,0.5,0.5}
\definecolor{codepurple}{rgb}{0.58,0,0.82}
\definecolor{backcolour}{rgb}{0.95,0.95,0.92}

\lstdefinestyle{pythonStyle}{
    commentstyle= \color{red!50!green!50!blue!50},		
    basicstyle=\ttfamily\footnotesize,
    breakatwhitespace=false,         
    breaklines=true,		
    captionpos=b,                    
    keepspaces=true,                 
    numbers=none,		
    numbersep=5pt,                  
    showspaces=false,                
    showstringspaces=false,		
    showtabs=false,                  
    tabsize=2,
    frame=single,	
}

\usepackage{makecell}

\begin{document}

\title{EduNLP: Towards a Unified and Modularized Library for Educational Resources}

\author{Zhenya Huang}
\affiliation{%
  \institution{University of Science and Technology of China}
  \city{Hefei}
  \country{China}
}
\email{huangzhy@ustc.edu.cn}

\author{Yuting Ning}
\authornote{Both authors contributed equally to this research.}
\affiliation{%
  \institution{University of Science and Technology of China}
  \city{Hefei}
  \country{China}
}
\email{ningyt@mail.ustc.edu.cn}

\author{Longhu Qin}
\authornotemark[1]
\affiliation{%
  \institution{University of Science and Technology of China}
  \city{Hefei}
  \country{China}
}
\email{qlonghu@mail.ustc.edu.cn}

\author{Shiwei Tong}
\affiliation{%
 \institution{Tencent Company}
 \city{Shenzhen}
 \country{China}
}
\email{shiweitong@tencent.com}

\author{Shangzi Xue}
\affiliation{%
  \institution{University of Science and Technology of China}
  \city{Hefei}
  \country{China}
}
\email{xueshangzi@mail.ustc.edu.cn}

\author{Tong Xiao}
\affiliation{%
  \institution{University of Science and Technology of China}
  \city{Hefei}
  \country{China}
}
\email{tongxiao2002@mail.ustc.edu.cn}

\author{Xin Lin}
\affiliation{%
  \institution{University of Science and Technology of China}
  \city{Hefei}
  \country{China}
}
\email{linx@mail.ustc.edu.cn}

\author{Jiayu Liu}
\affiliation{%
  \institution{University of Science and Technology of China}
  \city{Hefei}
  \country{China}
}
\email{jy251198@mail.ustc.edu.cn}

\author{Qi Liu}
\authornote{Corresponding author.}
\affiliation{%
  \institution{University of Science and Technology of China}
  \city{Hefei}
  \country{China}
}
\email{qiliuql@ustc.edu.cn}

\author{Enhong Chen}
\affiliation{%
  \institution{University of Science and Technology of China}
  \city{Hefei}
  \country{China}
}
\email{cheneh@ustc.edu.cn}

\author{Shijin Wang}
\affiliation{%
  \institution{
iFLYTEK Research}
  \city{Hefei}
  \country{China}
}
\email{sjwang3@iflytek.com}

\renewcommand{\shortauthors}{Huang, Ning and Qin, et al.}

\begin{abstract}
  Educational resource understanding is vital to online learning platforms, which have demonstrated growing applications recently. However, researchers and developers always struggle with using existing general natural language toolkits or domain-specific models. The issue raises a need to develop an effective and easy-to-use one that benefits AI education-related research and applications. To bridge this gap, we present a unified, modularized, and extensive library, \textit{EduNLP}, focusing on educational resource understanding.
  In the library, we decouple the whole workflow to four key modules with consistent interfaces including data configuration, processing, model implementation, and model evaluation. We also provide a configurable pipeline to unify the data usage and model usage in standard ways, where users can customize their own needs. For the current version, we primarily provide 10 typical models from four categories, and 5 common downstream-evaluation tasks in the education domain on 8 subjects for users' usage. The project is released at: https://github.com/bigdata-ustc/EduNLP.

\end{abstract}



\received{20 February 2007}
\received[revised]{12 March 2009}
\received[accepted]{5 June 2009}

\maketitle

\section{Introduction}

\begin{figure*}
    \centering
    \includegraphics[width=0.8\textwidth]{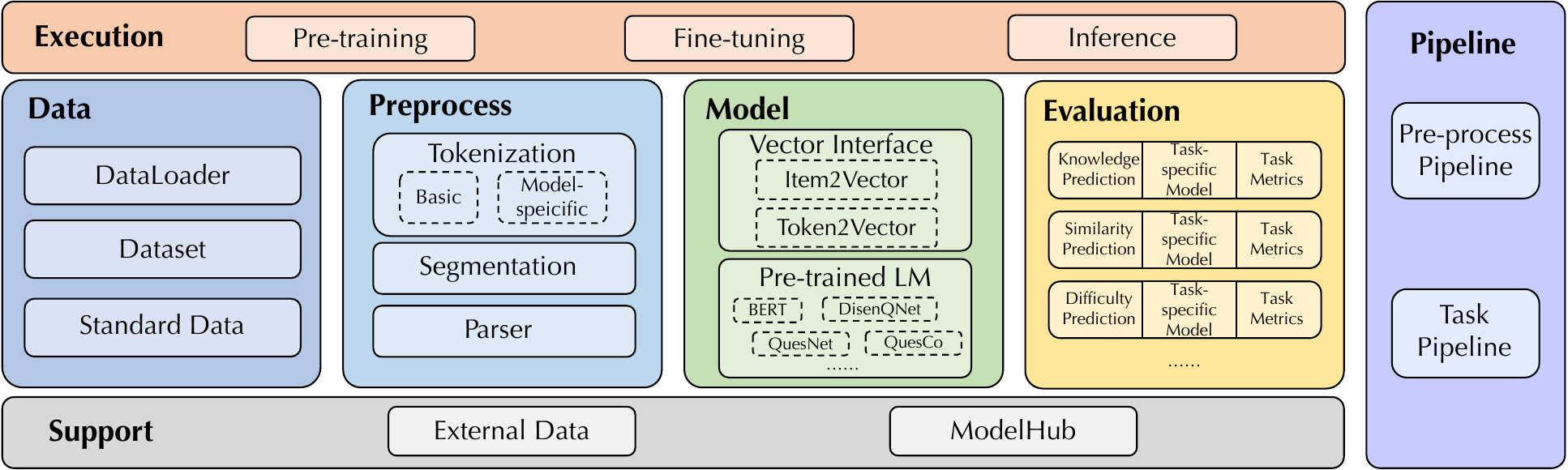}
    \vspace{-3mm}
    \caption{The overall framework of our library EduNLP.}
    \vspace{-5mm}
    \label{fig:framework}
\end{figure*}


Online learning platforms have been increasingly popular worldwide in recent years, which collect and manage large volumes of different educational resources (e.g., questions and textbooks) to provide personalized services for learners, such as learning path planning and personalized exercise recommendation~\cite{wu2020exercise, liu2019ekt, wang2022textbook}. For example, Khan Academy\footnote{https://www.khanacademy.org/} has collected more than 1 million exercises and served millions of users\footnote{https://www.librarieswithoutborders.org/khan-academy/}. To improve the quality of intelligent services, it is necessary to understand and analyze these educational resources to make the best of them~\cite{ fang2021guided, tlili2021towards, yin2019quesnet}. Such demands promote growing attention in many awesome intelligent education products. For example, OpenAI has provided a guide about using their awesome ChatGPT in teaching\footnote{https://openai.com/blog/teaching-with-ai}. Therefore, there is a need for researchers and developers from both academia and industry to incorporate cutting-edge data mining and AI techniques to promote both research and applications of AI in education~\cite{wang2022towards, gan2023large, lu2023readingquizmaker}.

Although there are already mature NLP toolkits or even promising LLMs (e.g., ChatGPT) that we could use~\cite{wang2023hug, Gardner2017AllenNLP, li2021textbox, wolf2019huggingface}, they generally do not work well on such educational resources due to their unique characteristics. First, educational resources usually contain multiple necessary components besides texts. For example, as shown in Figure~\ref{fig:sif}, the typical math questions consist of texts, formulas, images and additional side information (such as related knowledge concepts)~\cite{chen2021geoqa, lu2023mathvista}. Second, different from the common corpus, educational resources are always coupled with much domain-specific knowledge and logic, which is crucial for analyses. For instance, to understand and reason on math questions, models are required to leverage complex math knowledge\cite{liu2023learning, lin2021hms, liu2023guiding, liu2022cognitive}. Directly adopting NLP models for semantic understanding is insufficient, as many logical features are overlooked~\cite{yin2019quesnet, ning2023towards}.

Recently, researchers have noticed the above issues and designed many domain-specific models for educational resource understanding, such as QuesNet~\cite{yin2019quesnet}, DisenQet~\cite{huang2021disenqnet}, QuesCo~\cite{ning2023towards}, and JiuZhang~\cite{zhao2022jiuzhang, zhao2023jiuzhang2}. However, their implementations are usually tough to use, which brings obstacles to practical applications and further research. Even experienced researchers also struggle with the following two main issues. First, these implementations usually adopt inconsistent data formats and model interfaces, which makes it hard to reproduce across different models and datasets. Second, the modules (e.g., data preprocessing, model details) in the implementations are tightly coupled, which increases the difficulty of secondary development and further extension. Therefore, researchers/developers are eagerly seeking an easy-to-use toolkit to improve their efficiency, which raises a need to re-implement these methods in a unified way, benefiting the related fields.

To this end, we initiate the \textit{EduNLP} project to provide a unified and modularized library for education resource understanding. We decouple the typical workflow, including data configuration, preprocessing, model implementation, and model evaluation, into four modules with unified interfaces. Based on this library, we would like to offer researchers and developers two key conveniences including easy-to-use reproducing and customized secondary development.
Specifically, EduNLP has the following key features.



\begin{itemize}
    \item EduNLP is a modularized framework. It decouples the whole workflow of model development into different modules including data configuration, data processing, model training, and model inference. We also encapsulate all modules and provide an easy-to-use pipeline interface to meet a variety of personalized needs of users, such as quick inference for downstream tasks, configurable preprocessing, etc.
    \item EduNLP is compatible with typical educational resource data. We design a standard item format (SIF) to standardize the common features in the data including text, formula, image, and attributes (e.g., knowledge concept label), etc. Based on SIF, we also provide useful preprocessing tools that support users in processing their data with parsing, segmentation, tokenization, etc, which could reduce the redundancy for development.
    \item EduNLP unifies the model development process. We design an abstract class as the base model for users to implement their own ones. We also provide vector interfaces for easy inference. Users can directly download the models in our library for usage, or pre-train/fine-tune models with new data on their own, etc.
    \item EduNLP is equipped with several common downstream tasks, facilitating fair evaluation. Users are also supported to develop their application tasks for standard evaluations.
    \item Primarily, we implement 10 typical models in our library by default with four categories including embedding models, Seq2Seq models, general pre-trained models, and educational pre-trained models. We also provide 5 common downstream evaluation tasks including knowledge prediction, difficulty prediction, etc, on 8 subjects (e.g., math, English). 
    \item EduNLP is flexible and extensible. It is based on PyTorch~\cite{paszke2019pytorch} and is compatible with HuggingFace~\cite{wolf2019huggingface}, which are two of the most popular frameworks in the NLP community, allowing researchers and developers to quickly adapt to our library and integrate new models into our library.

\end{itemize}

\section{The Library - EduNLP}
The overall framework of our library EduNLP is presented in Figure~\ref{fig:framework}. The data, preprocess, model, and evaluation modules are the core components of our library, which compose the whole workflow from data input to model output. Based on the four core modules, we provide the scripts to pre-train, fine-tune, and evaluate each implemented model. We also design a pipeline module for facilitating pre-processing data and running existing models on downstream applications. Besides, our library also supports downloading pre-trained models with a Model Hub and using external data.
In the following, we will introduce each module in detail.

\subsection{Data Module}
A major guiding principle in the development of our toolkit is to provide standardized and procedural data processing workflows for education resources.
For reusability and extensibility, we design a standard data format for educational resources, and provide an unified data flow to feed the input data into models. Compatible with classical practice in DL community~\cite{paszke2019pytorch, tensorflow2015-whitepaper, chollet2015keras}, our data flow can be described as as: input data $\rightarrow$ \textit{Dataset} $\rightarrow$ \textit{DataLoader} $\rightarrow$ models.

\subsubsection{Standard Data} \label{s211}

\begin{figure}[t]
    \centering
    \includegraphics[width=0.85\columnwidth]{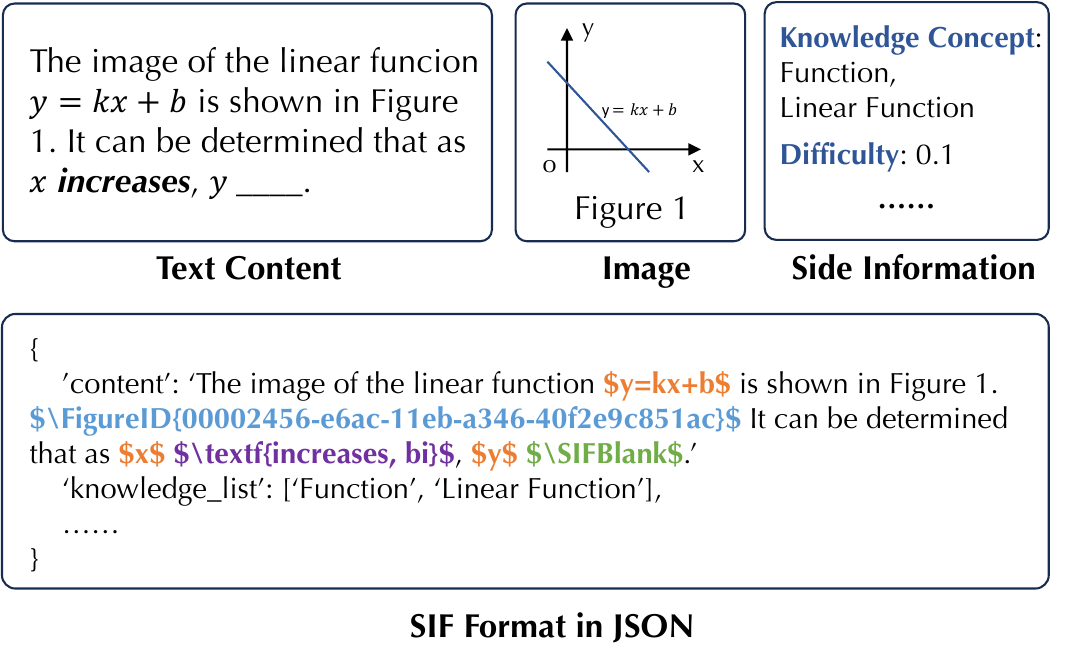}
    \vspace{-5mm}
    \caption{An example of SIF item.}
    \vspace{-6mm}
    \label{fig:sif}
\end{figure}

An educational item (such as a test question or a textbook snippet) may contain complex components represented in various formats. For example, in Figure~\ref{fig:sif}, the question is composed of normal texts (``The image ... function''), formulas (``$y=kx+b$'' in text content), special tokens (``\_\_'' for blank to fill in text), images (``Figure 1'') and side information (``Knowledge Concept'' and ``Difficulty''). Besides, some special words in text may be displayed in different styles (e.g., italic or bold).
The complexity of the educational item leads to an inconsistent data format from different data sources, 
which presents difficulties in reusable data pre-processing and creating unified model structures. 

Therefore, to facilitate future research and development, we define a standard item format (SIF) for educational resources compatible with most components. An example of the SIF format is also given in the bottom of Figure~\ref{fig:sif}.

The main principles of our SIF format are as follows:
\begin{itemize}
    \item \textbf{Normal Text:} Currently we only support Chinese and English. Only characters and punctuation in Chinese and English, as well as line breaks, are allowed.
    
    \item \textbf{Formulas: } Mathematical symbols like formulas, Roman characters and numbers need to be expressed in \LaTeX \footnote{https://www.latex-project.org/} format, i.e., embedded as \texttt{\$...\$}.
    
    \item \textbf{Special tokens:} Represent underlines of blanks and brackets of choices with \texttt{\$\textbackslash SIFBlank\$} and \texttt{\$\textbackslash SIFChoice\$} respectively. For multi-choice questions, the options should be started with \texttt{\$\textbackslash SIFTag\{options\}\$} and separated by \texttt{\$\textbackslash SIFSep\$}.
    
    \item \textbf{Text styles:} Text can be represented in different styles with \texttt{\$\textbackslash textf\{text, style\_labels\}\$}. Currently, we have defined some styles: b(bold), i(italic), u(underline), w(wave), d(dotted), t(title). The parameter \texttt{style\_labels} can be a combination with these styles in alphabetical order.
    
    \item \textbf{Images:} 
    Pictures and tables can be represented as Base64 or $ FigureID \{ uuid \} $. 
    Especially,\texttt{ \$\textbackslash FormFigureID\{uuid\}\$} is used to represent formula pictures,
    which can be further converted into latex formula 
    LaTeX strings using our OCR module base on GAP\cite{yang2024gap} for multi-line formula recognition. 
    
    \item \textbf{Side information:} Side information (such as knowledge concepts and difficulty), if any, can be additionally represented separately.
    
\end{itemize}

\subsubsection{Dataset and Dataloader}\label{s212}

To facilitate preparing data for model training, we develop the \texttt{EduDataset} as the container for the data, which automatically conducts data preprocessing to convert the input data into the form required by the model, including parsing, segmentation and tokenization (detailed in Section~\ref{s22}). Notably, \texttt{EduDataset} also supports multi-process acceleration for dataset processing.
Following classical practice in DL community~\cite{paszke2019pytorch, tensorflow2015-whitepaper, chollet2015keras}, we also develop the \texttt{Dataloader} class based on \texttt{Dataset} to organize the data stream in model training, such as sampling data and collating them into a batch.

\subsubsection{External Data}
EduNLP allows usage of any external data collected by users or downloaded from existing data sources, as long as they are processed into SIF format, such as mathematical problems, reading comprehension, and essay questions.
Moreover, to facilitate the use of educational resources data for developers, we collected some open-source datasets of test questions, including mathematical word problems~\cite{math23k-wang-etal-2017-deep,  Ape210k-zhao2020ape210k, SVAMP-patel-etal-2021-nlp, MaWPS-koncel-kedziorski-etal-2016-mawps,GSM8K-cobbe2021training, AQUA-RAT-ling-etal-2017-program}, equation problems~\cite{Hmwp-qin-etal-2020-semantically, Lmwp-liu-2021-make, mathQA-amini-etal-2019-mathqa} and so on. We have made these open datasets available with a data tool EduData\footnote{https://github.com/bigdata-ustc/EduData}, which provides a convenient downloading interface for researchers to access the data. Specifically, users can directly obtain the data through the command \texttt{edudata download DATASET\_NAME}.

\subsection{Preprocess Module} \label{s22}
When feeding the data into models, we need to pre-process the data, i.e., converting an educational item into a sequence of token ids. Considering the complex components in educational items, in EduNLP, we design the workflow of data processing as: Item Format Parse $\rightarrow$ Component Segmentation $\rightarrow$ Tokenization.

\subsubsection{Item Format Parse}
The module \texttt{Parse} is the foundation of the SIF format introduced in Section~\ref{s211}, which identifies different components in the text (e.g., letters, numbers, formulas, and math symbols) and converts them into SIF format. Specifically, this module implements essential format parsing tools including identifying Chinese characters, detecting English alphabets and numerals outside mathematical expressions, and converting special elements (e.g., blanks (`\_\_') and parentheses (`()')) into predefined standard symbols (e.g., \texttt{\$\textbackslash SIFChoice\$}, \texttt{\$\textbackslash SIFBlank\$}).

For the convenience of the users, we provide two easy-to-use interfaces containing all the above processes:
\begin{itemize}
\item \texttt{is\_sif()}: This function checks if the raw data adheres to the SIF format, allowing developers to quickly identify conversion requirements for compatibility with our library.
\item \texttt{to\_sif()}: This function converts non-SIF data into SIF format, simplifying preprocessing by eliminating developers' concerns over original data formats.
\end{itemize}

Furthermore, we also offer more advanced features including matching \LaTeX~mathematical expressions, verifying their completeness, converting image-based formulas into \LaTeX~format, and encoding images in Base64 format.

\subsubsection{Component Segmentation}

As we mentioned before, educational resources usually contain multiple components besides text, such as formulas and images. Therefore, this module aims to identify these components and decompose the educational item into different segments based on the SIF format.

Specifically, given a SIF item, the segmentation involves identifying and categorizing various components (including plain text, mathematical expressions and graphical images and so on) in the content. We define special data structures for each component, i.e., \texttt{TextSegment}, \texttt{FormulaSegment}, \texttt{FigureSegment}, \texttt{FigureFormul-\\aSegment} and \texttt{TagSegment}.

We provide an easy-to-use interface, \texttt{seg()}, for the segmentation process, which will yield a \texttt{SegmentList} object as a list of all the components. Besides, users can use the parameter \texttt{symbol} of the function to mask specific components. For example, if one would like to focus on pure text, he can set \texttt{symbol} to \texttt{`fgm'} (i.e., formulas, figures and marks), and the specified components will be masked.

\subsubsection{Tokenization}
After getting the components in an educational item, this module tokenizes each component into a serialized sequence of tokens required by the models. The tokens can be pre-defined formula tokens, text tokens, and image tokens, etc.

\textbf{Text tokenization} typically refers to word segmentation, i.e., transforming a sentence into a sequence of words. In EduNLP, we provide multiple candidate tokenization methods, such as Byte Pair Encoding (BPE) based on Huggingface Tokenizers\footnote{https://huggingface.co/docs/tokenizers}, the Jieba Chinese tokenization algorithm\footnote{https://github.com/fxsjy/jieba}, and nltk tokenization\footnote{https://www.nltk.org/}. Notably, we develop an \texttt{EduVocab} class to hold the vocabulary for educational resources tokenization.

\textbf{Formulas tokenization} is provided with two typical implementations, i.e., the linear parsing which only keeps the symbols in the formula, and the abstract syntax tree parsing which also keeps the structural information. 

Linear parsing of formulas provides a standardized  tokenization method for \LaTeX~formulas. This is crucial because naive text-based serialization can inadvertently introduce information redundancy or loss. For instance, the fraction command `\textbackslash frac' might be incorrectly tokenized into `\textbackslash' and `\texttt{frac}'. We have designed specialized formula rules to identify and process specific identifiers in \LaTeX, such as `\{\}', `\textbackslash frac', and `\textbackslash circ'. Abstract syntax tree (AST) parsing is implemented by the \texttt{Formula} class, which checks whether formulas are valid and parses the valid ones to AST. To yield a structure-enhanced formula token sequence, we provide the \texttt{ast\_tokenize} function to serialize the nodes.

Moreover, we provide the \texttt{FormulaGroup} class to parse multiple formulas into an AST forest.
After acquiring the AST tree or forest, following~\cite{huang2020neural}, we further construct a formula graph by adding several relation edges. Based on the formula graph, EduNLP could support more complex modeling methods such as the graph neural network, which we plan to add in the future.

\textbf{Images tokenization} refers to instantiating image with base64 encodings or reading image from the related path with its UUID, which will convert images into tensors in models supporting image modeling such as QuesNet~\cite{yin2019quesnet}.

    

\subsubsection{Tokenizer}
\label{s224}

To facilitate the standardized data processing and simplify usage, we design an interface \texttt{Tokenizer} that integrates \texttt{Parse}, \texttt{Segmentation} and \texttt{Tokenization} altogether. \texttt{Tokenizer} is called explicitly by the user or implicitly by \texttt{EduDataset} (see Section~\ref{s212}) to pre-process data for models. 

We first provide two pre-packaged basic tokenizers:
\begin{itemize} 
    \item \texttt{PureTextTokenizer} processes the text in the item by segmenting words and applies linear parsing methods to formulas in textual format, while other elements such as images and tags are symbolized (e.g., replace the image with a special \texttt{[FIGURE]} token). It matches situations where only the text contents are required.
    \item \texttt{AstFormulaTokenizer} is similar to \texttt{PureTextTokenizer}, but employs structural tokenization based on abstract syntax tree parsing for formulas. It works in situations where the structural information of the formulas is non-trivial.
\end{itemize}
We also provide the \texttt{CustomTokenizer} class, enabling users to specify configurations for text, formula, and image processing on their demands.

Furthermore, building on basic tokenizers, we design several model-specific tokenizers to facilitate model training and inference. The model-specific tokenizers follow two special class structures:
\begin{itemize}
    \item \texttt{TokenizerForHuggingface} is an abstract class designed to reuse Huggingface's tokenizers. For example, our implemented \texttt{BertTokenizer} replaces the basic tokenization \texttt{SentencePiece} in \texttt{transformers.BertTokenizer} with our basic tokenizers \texttt{PureTextTokenizer} or \texttt{AstFormulaToken-\\izer}. These tokenizers work for most general pre-trained language models.
    \item \texttt{PretrainedEduTokenizer} is an abstract class for all the customized tokenizers including ElmoTokenizer, DisenQTokenizer, QuesNetTokenizer and so on, which typically work for models specially designed for educational resources like DisenQNet and QuesNet. Users can also implement their customized tokenizers specially tailored to their own models based on this class.
\end{itemize}



\subsection{Model Module}

In this module, we organize several categories of models into unified implementations. We first design an abstract model that specifies several important interfaces, and then implement 4 categories of models based on the abstract model. For all implemented models, users can directly download and use our pre-trained models from the Model Hub, or pre-train their own model with our provided pre-training scripts.

\subsubsection{Unified Interface Design}
To unify the implementation and usage of various models, we design an abstract model, i.e., \texttt{BaseModel}, that defines the necessary interfaces to work with other components. Therefore, based on the abstract model, all the models can be used with unified interfaces and easily connected to other components in our library. The abstract model is also compatible with HuggingFace Transformers~\cite{wolf2019huggingface}, which allows direct adaption from a large number of models in HuggingFace to our library.

Specifically, compatible with the PyTorch and HuggingFace community, to implement a model, users should at least implement the \texttt{\_\_init\_\_()} function to define the model architecture and configurations, the \texttt{forward()} function to compute the loss and embeddings, the \texttt{save\_config()} function to save the configuration file and the \texttt{from\_config()} to initialize the model from an existing configuration file. More details and interfaces will be introduced in Section~\ref{subsec:new_model}. 
Through these general interfaces for representation models, users can implement various models in a highly unified way, and directly reuse other components in our library with their new models, such as pipelines.

\subsubsection{Implemented Models}
\label{sssec:implemented_models}
We have implemented 10 models inherited from our abstract model \texttt{BaseModel} until now. The implemented models cover the general representation methods commonly used in the past years, and some representative language models specially designed for educational resources. We roughly divided these models into 4 categories: word embedding models, seq2seq models, general pre-trained language models, and educational pre-trained language models. Table~\ref{tab:models} shows the details of these models.

\begin{table}[t]
    \centering
    \caption{We implemented 10 models in EduNLP.}
    \vspace{-3mm}
    \small
    \begin{tabular}{ccc}
    \toprule
       \textbf{Category}  & \textbf{Model} & \textbf{Year} \\
       \midrule
        \multirow{2}{*}{Word Embedding} & Word2Vec~\cite{mikolov2013efficient} & 2013 \\
        & Doc2Vec~\cite{le2014distributed} & 2014 \\
        \midrule
        \multirow{1}{*}{Seq2Seq Model} & Elmo~\cite{peters-etal-2018-deep} & 2018\\
        \midrule
        \multirow{3}{*}{\makecell[cc]{General Pre-trained \\ Language Model}} & BERT~\cite{devlin-etal-2019-bert} & 2019\\
        & RoBERTa~\cite{liu2019roberta} & 2019 \\
        & MacBERT~\cite{cui2020revisiting} & 2020 \\
        \midrule
        \multirow{4}{*}{\makecell[cc]{Educational Pre-trained \\ Language Model}} & QuesNet~\cite{yin2019quesnet} & 2019 \\
        & DisenQNet~\cite{huang2021disenqnet} & 2021\\
        & JiuZhang~\cite{zhao2022jiuzhang} & 2022 \\
        & QuesCo~\cite{ning2023towards} & 2023 \\
    \bottomrule
    \end{tabular}
    \label{tab:models}
    \vspace{-6mm}
\end{table}

For each implemented model, we provide the pre-training script in EduNLP (e.g., the \texttt{pretrain\_disenq()} for pre-training DisenQNet), and pre-train the model on our large-scale educational dataset. We also test the performance of these models on 5 downstream tasks with our multi-subject educational dataset (Section ~\ref{sec:exp}).

\subsubsection{Vector Interfaces}
\label{sssec:container}
To facilitate the inference of implemented models (i.e., converting educational items like test problems into vectors with the model), we provide two types of API interfaces for them, i.e. Item2Vector (\texttt{I2V}) and Token2Vector (\texttt{T2V}). 
Specifically, Token2Vector performs model inference to convert the tokenized items (e.g., token list) into embedding vectors. Item2Vector further integrates the data pre-processing and model inference, and directly converts the input educational items (e.g., test problems) into output vectors. The two interfaces both support using a local model or loading our pre-trained models remotely. Section~\ref{sssec:inference} shows the examples of inference with Item2Vector and Token2Vector.

\subsubsection{Model Hub}
To facilitate easy use of these models with EduNLP, following HuggingFace Transformers ~\cite{wolf2019huggingface}, we build a Model Hub\footnote{https://modelhub.bdaa.pro/} to save and distribute our pre-trained models. We have uploaded the models pre-trained and fine-tuned on our educational data to Model Hub. We organize models with collections. Specifically, models with the same structure but different data or configurations can be uploaded in one collection for better management. 



To download a model from the Model Hub, users can download one from the website or with our APIs (e.g., download a model with \texttt{/api/getPretrainedModel}).
They can also download a model and load it with I2V container by the \texttt{get\_pretrained\_i2v()} function.

\subsection{Evaluation Module}
\label{subsec:eval}
It is important to evaluate model output (e.g., the embedding vectors generated by the models) with unified downstream tasks. Therefore, EduNLP provides several common downstream tasks in real intelligent education scenarios~\cite{huang2019hierarchical,  huang2017question, byrd-srivastava-2022-predicting, liu2018finding} for fair evaluation, i.e., knowledge prediction, difficulty prediction, discrimination prediction, quality prediction and similarity prediction.


\begin{itemize}
    \item \textbf{Knowledge prediction} is a multi-label classification task. Given an educational item (such as a test question), the task aims to predict the related knowledge concepts from a predefined concept pool.
    \item \textbf{Difficulty prediction} is a regression task, which aims to quantify the degree of the difficulty of a given question. The difficulty score is defined on a scale from 0 to 1, where a higher score indicates a more difficult question.
    \item \textbf{Discrimination prediction} is a regression task, which aims to estimate the discrimination of a given problem for students with varying proficiency levels. 
    A problem with good discriminatory score means that students with higher ability have a higher success rate on this question than those with lower ability.
    The discrimination prediction shares the same settings of difficulty prediction.
    \item \textbf{Quality prediction} aims to predict the quality of test questions. The quality scores of test questions are rated from 1 to 5. A high-quality score means that the question is presented in clear text content and standard formula formats and is of high logical consistency. It is worth noting that in this task, we need to first categorize the test questions into three scenarios (i.e., daily practice exercises, quizzes, and final exams) and then predict their quality scores. 
    \item \textbf{Similarity prediction} aims to predict the similarity between two given educational items. Here we use the cosine similarity of the pre-trained item representations without further fine-tuning for zero-shot evaluation.
\end{itemize}

Please refer to Appendix~\ref{appe:Downstream} for more detailed task descriptions. For each downstream task, we implement a task-specific model following the same design with our abstract model. The task-specific model contains a pre-trained language model as the encoder and some prediction layers. For each downstream task, EduNLP implements several common evaluation metrics (Table~\ref{tab:metrics}). We provide fine-tuning functions in EduNLP (e.g., the \texttt{finetune\_disenq\_for\_\\knowledge\_prediction()} function for DisenQNet on knowledge prediction task). We also provide scripts for fine-tuning and evaluating each task-specific model in \texttt{examples/downstream}.

\begin{table}[t]
    \centering
    \caption{We provide several evaluation metrics.}
    \vspace{-3mm}
    \small
    \begin{tabular}{l|l}
    \toprule
        \textbf{Task} &  \textbf{Evaluation Metrics} \\
        \midrule
        Knowledge Prediction & precision, recall, f1-score \\
        \midrule
        Difficulty Prediction & \makecell[tl]{MAE, MSE, RMSE, NDCG, \\Pearson correlation coefficient, \\Spearman correlation coefficient} \\
        \midrule
        Discrimination Prediction & \makecell[tl]{MAE, MSE, RMSE, NDCG, \\Pearson correlation coefficient, \\Spearman correlation coefficient}\\
        \midrule
        Quality Prediction & \makecell[tl]{accuracy, MSE, R2Score, \\Person correlation coefficient}\\
        \midrule
        Similarity Prediction & \makecell[tl]{Pearson correlation coefficient, \\Spearman correlation coefficient} \\
    \bottomrule
    \end{tabular}
    \vspace{-5mm}
    \label{tab:metrics}
\end{table}

\subsection{Pipeline Module}
To provide a unified and easy-to-use interface for downstream inference, we implement a configurable \texttt{pipeline} that integrates all components in our library to form the whole workflow from data pre-processing to downstream tasks.

\subsubsection{Task Pipeline}
Pipelines are easy ways to use pre-trained models in downstream tasks. To set a pipeline for the downstream task, users only need to specify the task, model and tokenizer in the \texttt{pipeline()} interface. Then for the input educational items, the pipeline will automatically process the input with the configured tokenizer, construct \texttt{Dataset} and \texttt{DataLoader}, load the task-specific model, and feed the data into the model to get the predictions for the specified tasks.

\subsubsection{Pre-process Pipeline}
More flexibly, for users who only want to use our library to process the educational data without models, 
the pipeline could also be configured for data pre-processing.
The general pre-processing process in our library consists of SIF format conversion, segmentation, and tokenization. Moreover, some of the steps may be not necessary, and sometimes there is a need to add more processing steps. 
To this end, EduNLP provides a configurable way to arrange the pre-processing steps of the pipeline.

Users can initialize a pre-processing pipeline with the \texttt{pipeline()} interface and add some initial steps in the meantime. After setting up a pipeline, users can still add new steps by \texttt{pipeline.add\_pipe()} interface and set the configuration for the steps.

We also support the combined configuration of the pipeline for pre-processing and downstream task. Users can use the parameter \texttt{task} to specify the downstream tasks and pass a list of pre-processing steps to the parameter \texttt{preprocess} in the same pipeline. The pipeline will first process the input data with the configured steps, and then run the specified downstream tasks to get the predictions.

\section{Library Usage}
\begin{figure}
    \centering
    \includegraphics[width=0.9\columnwidth]{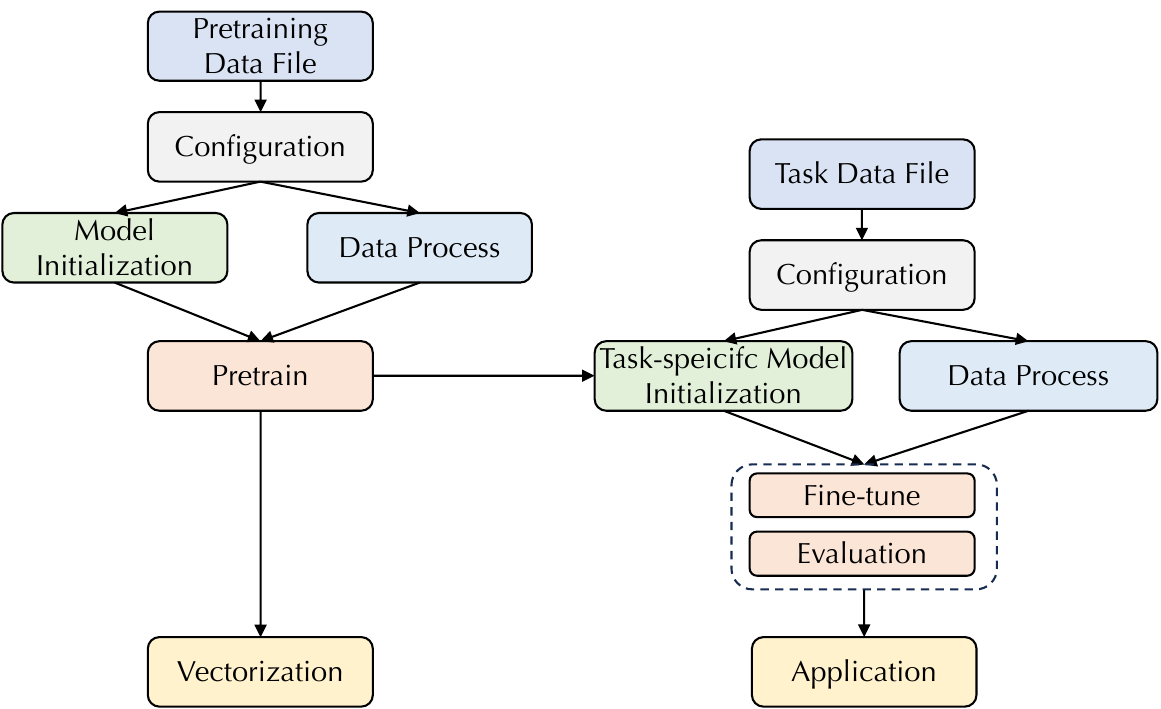}
    \vspace{-2mm}
    \caption{An illustrative usage flow of EduNLP.}
    \label{fig:usage}
\end{figure}

In this section, we introduce how to use our library. Figure~\ref{fig:usage} shows the general procedure of EduNLP. We will illustrate some key usage of our library in this section, i.e., how to prepare the dataset, how to run existing models and how to implement a new model quickly. We will take DisenQNet~\cite{huang2021disenqnet} as an example in the following examples. Please refer to our documentation~\footnote{https://edunlp.readthedocs.io/en/latest} for more detailed usage.

\subsection{Processing New Data}
\label{subsec:process_dataset}
Before training a model, it is essential to process the data and ensure it is in the appropriate format.

As mentioned in Section~\ref{s22}, we provide the \texttt{is\_sif()} and \texttt{to\_sif()} functions for users to first check whether their data is in SIF format and convert their raw data into SIF format if not. Then users can instantiate a tokenizer that conducts the whole data pre-processing (i.e., parsing, segmentation and tokenization) and converts tokenized items into ids as the inputs to models. One step further, the users should prepare the dataset based on the data and tokenizer with our \texttt{Dataset} for training.

Code~\ref{code:prepare_data} shows an example of preparing the data.

\begin{lstlisting}[language=Python, style=pythonStyle, label={code:prepare_data}, caption=A simple example for preparing data.]
from EduNLP.SIF import to_sif
# converting data to SIF format
sif_data = to_sif(raw_data)
# get a tokenizor
from EduNLP.Pretrain import DisenQTokenzier
tokenizer = DisenQTokenizer(vocab_path, text_params=text_params)
# prepare dataset
from EduNLP.Pretrain import DisenQDataset
dataset = DisenQDataset(sif_data, tokenizer=tokenizer)
\end{lstlisting}
\vspace{-2mm}

\subsection{Running Existing Models}
Users can directly load pre-trained model checkpoints and perform inference with them. Besides, EduNLP also supports users to pre-train and fine-tune the implemented models with their own data.

\subsubsection{Inference with a Pre-trained Model}
\label{sssec:inference}
To infer with an existing model, we provide an easy-to-use interface (i.e., Item2Vector (\texttt{I2V})) that can convert educational items (e.g., questions) to vectors. We provide an API to directly initialize an \texttt{I2V} instance with our pre-trained models, i.e., \texttt{get\_pretrained\_i2v(model\_name)}. It will download the model from our Model Hub and load the model with \texttt{I2V}. One can also load local models with model-specific \texttt{I2V} interfaces. After loading models with \texttt{I2V}, one can use the function \texttt{infer\_item\_vector()} to get the vectors of items and use the function \texttt{infer\_token\_vector()} to get the embeddings of all tokens in each item. 
Code~\ref{code:i2v} shows an example of using I2V interface for inference. We also provide another interface, Token2Vector (\texttt{T2V}), to convert tokenized items (i.e., tokens) to vectors, which is a subprocess of \texttt{I2V}.

\begin{lstlisting}[language=Python, style=pythonStyle, label={code:i2v}, caption=A simple example of using the I2V container.]
from EduNLP.I2V import DisenQ, get_pretrained_i2v
# load provided models from ModelHub directly
i2v = get_pretrained_i2v("disenq_test_128", model_dir=download_model_dir)
# load local models
tokenizer_kwargs = {"tokenizer_config_dir": model_dir}
i2v = DisenQ('disenq', 'disenq', model_dir, tokenizer_kwargs=tokenizer_kwargs)
# infer vectors
i_vec = i2v.infer_item_vector(items)
\end{lstlisting}

One step further, if users have their own data, they can also pre-train and fine-tune the models with their own data.

\subsubsection{Pre-training and Fine-tuning Models}

To pre-train an implemented model with EduNLP, users only need to process the dataset as we introduced in Section~\ref{subsec:process_dataset} and specify the training configurations. Then the pre-training can be run with simple interfaces in \texttt{EduNLP.Pretrain}. These functions will construct the dataset, initialize the model with specified configurations and then pre-train the model. The pre-training leverages the \texttt{Trainer} class in HuggingFace, supporting all the training features inherently. The pre-trained model will be saved to \texttt{output\_dir}. Code~\ref{code:pretrain} shows an example of pre-training a model. We provide code examples for pre-training all the implemented models in the folder \texttt{examples/pretrain} in EduNLP.

For downstream tasks, we have implemented five downstream tasks in EduNLP. Users can load the pre-trained language models and fine-tune them for a task-specific model. We implemented some interfaces to finetune the models, such as \texttt{finetune\_disenq\_for\_\\knowledge\_prediction()}. Code~\ref{code:pretrain} shows an example of fine-tuning a pre-trained DisenQNet for knowledge prediction. The fine-tuning also leverages the \texttt{Trainer} class in HuggingFace Transformers. One can also directly run the scripts in \texttt{examples/downstreams}.

\begin{lstlisting}[language=Python, style=pythonStyle, label={code:pretrain}, caption=A simple example for pre-training a DisenQNet model and fine-tuning it for knowledge prediction.]
from EduNLP.Pretrain import pretrain_disenqnet, finetune_disenqnet_for_knowledge_prediction
# pre-training
pretrain_disenqnet(train_items, output_dir,data_params=data_params, train_params=train_params)
# fine-tuning
finetune_disenqnet_for_knowledge_prediction(fine_items, task_output_dir, pretrained_model=output_dir, train_params=train_params, data_params=data_params)

\end{lstlisting}

\subsection{Implementing a New Model}
\label{subsec:new_model}

\begin{table*}[t]
\centering
\caption{Statistics for pre-training corpus on 8 subjects.}
\vspace{-3mm}
    \begin{tabular}{c|ccccccccc}
        \toprule
        \textbf{Subject} & English & Math    & Chinese  & Physics & Chemisty & Biology  & History & Geography \\
        \midrule
        \textbf{\#Problem}  & 403515   & 1495807 & 300336  & 889468  & 694154    & 443520    & 991212  & 218677   \\
        \bottomrule
    \end{tabular}
    \label{tab:data_pretrain}
\end{table*}

All the new models should inherit the \texttt{BaseModel} class defined in our library. It is worth noting that our \texttt{BaseModel} is compatible with HuggingFace Transformers, so that the user can also directly reuse a model or layer from HuggingFace. Based on our unified interface design, it is convenient to implement a new model by instantiating 4 mandatory functions as follows:
\begin{itemize}
    \item \texttt{\_\_init\_\_()} function. In this function, the user should perform parameter initialization, configuration setting, modules and layers definition and so on.
    \item \texttt{forward()} function. This function calculates the loss and item embedding. The user should first define a new output class, which should be a subclass of \texttt{ModelOutput}, for the new model. The \texttt{forward()} function should return an instance of the new output class.
    \item \texttt{save\_config()} function. This function saves the configuration files. It can be directly reused from the \texttt{BaseModel} if there is no special operations other than saving all the configuration parameters to one json file.
    \item \texttt{from\_config()} function. This function is a \texttt{classmethod} which initializes the new model with parameters from a configuration file.
\end{itemize}
We have also implemented the \texttt{save\_pretrained()} and \texttt{from\_pre-\\trained()} functions to save and load pre-trained checkpoints, which can be directly reused in general. One can also rewrite these two functions if needed.

In addition, to support easy inference, one can also implement the corresponding \texttt{I2V} (Item2Vector) and \texttt{T2V} (Token2Vector) class for the new model by re-written inference functions. More details are available in our documentation.

\section{Experiments}
\label{sec:exp}

To evaluate the models in EduNLP, we conduct extensive experiments to compare their performances on 8 subjects and 5 downstream tasks which are common in educational scenarios.

\begin{table*}[th]
    \centering
    \small
    \caption{Performance comparison between different models for five educational downstream tasks, i.e., difficulty prediction, knowledge concept prediction, discrimination prediction, similarity prediction, quality prediction and paper segmentation.}
    \vspace{-3mm}
    \begin{tabular}{cccccccccccccc}
    \toprule
      \multirow{2}{*}{\textbf{Model}} & \multicolumn{3}{c}{\textbf{Knowledge}} & \multicolumn{2}{c}{\textbf{Difficulty}} & \multicolumn{2}{c}{\textbf{Discrimination}} & \multicolumn{2}{c}{\textbf{Quality}} & \multicolumn{2}{c}{\textbf{Similarity}}\\
    \cmidrule(lr){2-4}\cmidrule(lr){5-6}\cmidrule(lr){7-8}\cmidrule(lr){9-10}\cmidrule(lr){11-12}
     & Precision ($\uparrow$) & Recall ($\uparrow$) & F1 ($\uparrow$) & RMSE ($\downarrow$)& NDCG@all ($\uparrow$) & MAE ($\downarrow$) & MSE ($\downarrow$) & ACC ($\uparrow$) & MSE ($\downarrow$) & PCC ($\uparrow$) & SCC ($\uparrow$)\\
    \midrule
    \textbf{Word2Vec}    & 0.5826 & 0.6597 & 0.6188 & 1.3844 & 0.8148          & 0.2032          & 0.0687          & 0.7013          & 0.1237          & 0.312           & 0.366           \\
\textbf{Doc2Vec}     & 0.5461 & 0.6426 & 0.5905 & 1.3417 & 0.8212          & 0.1735          & 0.0462          & 0.7234          & 0.2388          & 0.3487          & 0.2825          \\
\textbf{Elmo}        & 0.5397 & 0.6226 & 0.5782 & 1.4836 & 0.8142 & 0.2147          & 0.0875          & 0.7291          & 0.113           & 0.3557          & 0.3425          \\
\midrule
\textbf{Edu-BERT}    & \textbf{0.6197} & \textbf{0.7021} & \textbf{0.6583} & 1.1133 & 0.8241          & 0.1242          & 0.0248          & 0.7348          & 0.0582          & 0.6504          & 0.6407          \\
\textbf{Edu-MacBERT} & 0.6084 & 0.6997 & 0.6513 & \textbf{1.0993} & 0.8051          & 0.1245          & 0.0249          & 0.7242          & 0.0606          & 0.6189          & 0.6003          \\
\textbf{Edu-RoBERTa} & 0.6089  & 0.6986 & 0.6506 & 1.1041 & 0.8394          & 0.1241          & 0.0248          & 0.7283          & 0.0604          & 0.6269          & 0.6067          \\
\midrule
\textbf{QuesNet}     & 0.5964 & 0.6738 & 0.6328 & 1.3332 & 0.8102          & 0.1481          & 0.0432          & 0.7307 & 0.3349          & 0.3845          & 0.4516          \\
\textbf{DisenQNet}   & 0.5956 & 0.678  & 0.6342 & 1.5737 & 0.8244          & 0.1965          & 0.0453          & 0.7062          & 0.1209          & 0.475           & 0.4711          \\
\textbf{JiuZhang}    & 0.548  & 0.6486 & 0.5941 & 1.2788 & 0.8445          & 0.1203          & 0.0232          & 0.7242          & 0.0605          & 0.538           & 0.5402          \\
\textbf{QuesCo}      & 0.6036 & 0.6883 & 0.6432 & 1.1084 & \textbf{0.8737} & \textbf{0.1167} & \textbf{0.0221} & \textbf{0.7504} & \textbf{0.0567} & \textbf{0.7314} & \textbf{0.6986} \\
    \bottomrule
    \end{tabular}
    \label{tab:performance}
\end{table*}

\begin{table*}[th]
    \centering
    \caption{Model performance on 8 subjects and two tasks. We use F1-score and RMSE as the metrics for the knowledge prediction task and difficulty prediction tasks respectively. Note that `-' indicates DisenQNet lacks sufficient knowledge tags for pretraining.}
    \vspace{-3mm}
    \small
    \begin{tabular}{cccccccccc}
    \toprule
      \textbf{Task} & \textbf{Model} & \textbf{English} & \textbf{Math} & \textbf{Chinese} &  \textbf{History} & \textbf{Geometry} & 
      \textbf{Physics} &
      \textbf{Chemistry} & \textbf{Biology} \\
    \midrule
    \multirow{5}{*}{\textbf{\makecell[cc]{Knowledge \\ Prediction \\ (F1 $\uparrow$)}}} & \textbf{Word2Vec} & 0.5691& 0.6188 & 0.8735  & 0.5905 & 0.6399 & 0.7275 & 0.5693 & 0.7873 \\
    & \textbf{Doc2Vec} & 0.5757 & 0.5905 & 0.8886  & 0.5711 & 0.637 & 0.726 & 0.5603  & 0.7905 \\
    & \textbf{DisenQNet} & 0.6450 & 0.6342 & 0.5676  & - & -  & - &  -  &  -\\
    & \textbf{Edu-BERT} & \textbf{0.6914} & \textbf{0.6583} & \textbf{0.921}  & \textbf{0.7175} &  \textbf{0.7651} & \textbf{0.7882} &  \textbf{0.6477} &  \textbf{0.8117} \\
    & \textbf{Edu-RoBERTa} & 0.6638  & 0.6506 & 0.9151 & 0.7086 & 0.732 & 0.7723 & 0.6439 & \textbf{0.8117} \\
    \midrule
    \multirow{5}{*}{\textbf{\makecell[cc]{Difficulty \\ Prediction \\ (RMSE $\downarrow$)}}} & \textbf{Word2Vec}  & 1.2787 & 1.3844 & 0.9506 & 1.2365 & 1.1901 & 0.9484 & 0.9165 & 1.2671 \\
    & \textbf{Doc2Vec} & 1.2106 & 1.3417 & 0.9087 & 1.2142 & 1.1268 & 0.9024 & 0.8706 & 1.2488 \\
    & \textbf{DisenQNet} & \textbf{0.9628} & 1.5737 & 0.8822  & - & - & - & - & - \\
    & \textbf{Edu-BERT} & 0.9897 & 1.1133 & 0.8764  & \textbf{1.2088} & \textbf{0.9011} & \textbf{0.7496} & \textbf{0.7604} & \textbf{0.9958} \\
    & \textbf{Edu-RoBERTa} & 0.9759 & \textbf{1.1041} & \textbf{0.8042}  & 1.2492 & 1.0826 & 0.7556 & 0.9517 & 1.2462 \\     
    \bottomrule
    \end{tabular}
    \label{tab:multi-subjects}
\end{table*}

\subsection{Dataset and Setup}
\label{subsec:dataset}
We construct 8 large-scale educational datasets in different subjects as our pre-training corpus, consisting of millions of problems and papers from primary, junior high, and senior high school levels. The data is of high quality and in a unified format, including problem type, statement, answer, and analysis. Table~\ref{tab:data_pretrain} shows the detailed statistics with our data used in the library. Based on the datasets, we pre-train our implemented models on each subject. More training details can be found in Appendix~\ref{appe:Pretraining}.

For evaluation, we create a benchmark with 5 task-specific tagged datasets from real-world data. For example, the difficulty prediction and discrimination prediction dataset based on classical test theory~\cite{devellis2006classical} with actual student answer records. Please refer to Appendix~\ref{appe:Downstream} for the introduction of our datasets.


\subsection{Experimental Performance}
We conduct extensive experiments to test our models' performance and their domain generalization on diverse subject datasets.

\subsubsection{Performance on downstream tasks}

We use our implemented common educational tasks to evaluate the performances of all the implemented models in EduNLP. Due to the limited annotations of our dataset, we evaluate all the tasks in the math subject. Table~\ref{tab:performance} shows the performances of different models across 5 educational downstream tasks. For general language models, we further pre-trained them with the same educational pre-training dataset (i.e., getting Edu-BERT with BERT, Edu-MacBERT with MacBERT and Edu-RoBERTa with RoBERTa).

\subsubsection{Performance on Different Subjects}
As we mentioned in Section~\ref{subsec:dataset}, we pre-train models on 8 large-scale datasets of different subjects and get subject-specific pre-trained models. Since JiuZhang~\cite{zhao2022jiuzhang} and QuesCo~\cite{ning2023towards} are designed for educational math questions, we do not pre-train and evaluate them on other subjects. We selected some representative models in each category we introduced in Section~\ref{sssec:implemented_models} and compare them on 8 subjects with the knowledge prediction task and difficulty prediction task as examples. Table~\ref{tab:multi-subjects} presents the results.

\section{Discussion}
\label{sec:dis}
We developed EduNLP specifically for the processing of educational resources due to the inadequacies of existing toolkits, which can handle general textual data but fall short in dealing with typical educational data with diverse content components.
However, when large language models are increasingly dominating traditional NLP tasks, is it necessary to design such a sophisticated toolkit specially for educational resources?
Our answer is that it is still absolutely crucial and necessary for the following reasons.

First, high-quality data is essential for effective modeling, even for LLMs. EduNLP provided users with a standardized processing flow to handle high-quality data.
Second, domain-specific LLMs should be further trained with relevant tasks that consider domain characteristics\cite{gururangan2020don}, which is also a common solution for other resources such as code\cite{zheng2023codegeex, li2023starcoder, chen2023clex}. LMs usually perform better in educational tasks after trained in educational resources (Appendix~\ref{appe:domain-llm}).
As for general LLMs, they often struggle to achieve satisfactory results on educational tasks without fine-tuning(Appendix~\ref{appe:failure-llm}). 
Third, our library aims to provide a practical tool for researchers and developers, supporting them to start with educational resources easily. EduNLP is compatible with current popular frameworks (e.g. Huggingface Transformers). Users can quickly adapt to our toolkit and benefit from our features designed for educational resources.

\section{Conclusion}
In this paper, we introduced \textit{EduNLP}, a unified and modularized library for educational resource understanding to facilitate application and research in both industry and academics. EduNLP consists of four key modules covering the whole workflow (i.e., Data, Preprocess, Model, and Evaluation), with a unified interface for easy usage and secondary development. For data, we designed a standard item format for education resources and provided useful tools for data preprocessing. For model, we designed an abstract model as the interface, and implemented 10 pre-trained models of 4 categories for educational resources in the framework. We also evaluated these models with 5 unified downstream tasks in education scenarios.
In the future, we will continue expanding EduNLP with more pre-trained models. Moreover, we hope to incorporate the capability of educational resource generation in the next stage of EduNLP. 
We invite researchers and practitioners to join and enrich EduNLP, and help push forward the research on educational resource analysis.



\bibliographystyle{ACM-Reference-Format}
\bibliography{sample-base}

\appendix

\section{Corpus Examples} \label{appe:SIF}

Our pre-training datasets are extensive collections of practice quizzes, mock exams, and test problems. These data have been sourced from various educational supplementary materials providers. 
After data cleaning like filtering out HTML tags, unifying data structure and standardizing data formatting, we obtain the SIF datasets. Table ~\ref{tab:app_sif_sample} showcases illustrative examples of the SIF questions in different subject.

\section{Training Settings}  \label{appe:Pretraining}
During pre-training, we utilize the AdamW optimizer as the common optimizer for all models. We pre-train subject-specific models for each subject. Specifically, for RNN-based models, we pre-train them with a batch size of 64 and a learning rate of ${10}^{-3}$. For transformer-based models, we generally use a batch size of 32 and a learning rate of ${10}^{-5}$. We report the specific configurations we used for pre-training in our experiments in Table~\ref{tab:app_exp_settings}

\begin{table}[h]
    \centering
    \caption{Parameter settings of the baselines.}
    \vspace{-3mm}
    \begin{tabular}{l|l}
    \toprule
        \textbf{Model} &  \textbf{Training Settings} \\
        \midrule
        Word2Vec \& Doc2Vec & \makecell[tl]{window=5, min\_count=20\\ sg=1(W2V), dm=1(D2V), epoch=10 } \\
        \midrule
        Elmo & \makecell[tl]{AdamW, learning\_rate=$5e^{-3}$ \\batch\_size=64, epoch=5} \\
        \midrule
        \makecell[tl]{BERT \& RoBERTa \\\& MacBERT}  & \makecell[tl]{AdamW, learning\_rate=$5e^{-5}$ \\batch\_size=32, epoch=5 } \\
        \midrule
        QuesNet \& DisenQNet  & \makecell[tl]{AdamW, learning\_rate=$5e^{-4}$ \\batch\_size=64, epoch=10 } \\
        \midrule
        Jiuzhang & \makecell[tl]{AdamW, learning\_rate=$3e^{-5}$ \\batch\_size=32, \\epoch=2(stage1), 1(stage2), 1(stage3) } \\
        \midrule
        QuesCo & \makecell[tl]{AdamW, learning\_rate=$5e^{-5}$ \\batch\_size=8, epoch=1 } \\
    \bottomrule
    \end{tabular}
    \vspace{-2mm}
    \label{tab:app_exp_settings}
\end{table}


\section{Discussion about LLMs}  \label{appe:LLM}

\subsection{Effects of domain-specific training} \label{appe:domain-llm}

As we mentioned in Section~\ref{sec:dis}, LMs usually perform better in educational tasks after trained in educational resources. To prove it, we take BERT as an example, comparing the performance of original LMs and Edu-LMs that are further pre-trained on our SIF data in the similarity prediction task. Table~\ref{tab:app-domain-bert} shows that Edu-BERT performs much better than the original BERT, indicating the domain-specific training is important for LMs used in educational tasks. 

\begin{table}[h]
    \centering
    \caption{Edu-BERT that are further pre-trained on our SIF data perform much better than the original BERT in the similarity prediction task.}
    \begin{tabular}{ccc}
    \toprule
        \textbf{Metrics} & \textbf{BERT} & \textbf{Edu-BERT} \\
        \midrule
        PCC & 0.3694 & 0.6504\\
        SCC & 0.4133 & 0.6407 \\
    \bottomrule
    \end{tabular}
    \label{tab:app-domain-bert}
\end{table} 

\subsection{Failure cases of LLMs in educational tasks} \label{appe:failure-llm}
As we mentioned in Section~\ref{sec:dis},
We also tried prompting-based LLMs, but they did not achieve satisfactory results. Instances of ChatGPT are shown in Table \ref{tab:app_llm_bad_cases}. Specially, for knowledge prediction(KP), LLMs struggle to precisely predict knowledge labels due to output variability and lack of knowledge hierarchical information. For difficulty prediction(DP), LLMs demonstrate defect in perceiving difficulty compared to humans, even with contextual examples.

\begin{table}[h]
\setlength{\abovecaptionskip}{1pt}
\caption{Examples of bad cases of ChatGPT on two typical educational tasks.}
\label{tab:app_llm_bad_cases}
\centering
\resizebox{\linewidth}{!}{
\begin{tabular}{|p{8cm}|}
\hline
\centerline{\textbf{Knowledge Prediction}}
\textbf{\#Input}: \begin{CJK}{UTF8}{gbsn}设数列$a_{n}$的前n项和为$S_{n}$,若$S_{n}=2a_{n}-2^{n+1}(n \in N_{+})$,则数列$a_{n}$的通项公式为\$\textbackslash SIFBlank\$.\end{CJK}\newline
\textbf{\#Label}: 
\begin{CJK}{UTF8}{gbsn}[数列, 数列的概念和基本表示, 数列的递归关系]\end{CJK}
\newline
\textbf{\#Prediction}: 
\begin{CJK}{UTF8}{gbsn}[数列的通项公式, 数列的前n项和公式, 求和公式的运用]\end{CJK}
\\
\hline
\centerline{\textbf{Difficulty Prediction}}
\textbf{\#Input}: \begin{CJK}{UTF8}{gbsn}学校周三要排语文、数学、英语、物理、化学和生物6门不同的课程,若第一节不排语文且第六节排生物,则不同的排法共有多少种?\end{CJK} \newline
\textbf{\#Label}: 0.14 \newline
\textbf{\#Prediction}: 0.75 \\
\hline
\end{tabular}}
\end{table}

\section{Downstream Tasks}  \label{appe:Downstream} 


Our evaluation consists of two evaluation protocols, namely freezing evaluation and unfreezing evaluation. Specifically, freezing means that we freeze the language model as the encoder and only train the prediction networks, while unfreezing evaluation means that we will fine-tune the language model as well as the prediction networks. In our experiments, we adopt freezing evaluation for difficulty prediction, discrimination prediction and similarity prediction, and unfreezing evaluation for knowledge prediction and quality prediction.

\begin{table}[h]
\setlength{\abovecaptionskip}{1pt}
\caption{Examples of data samples of downstream tasks.}
\label{tab:app_downstream_sample}
\centering
\resizebox{\linewidth}{!}{
\begin{tabular}{|p{8cm}|}
\hline
\centerline{\textbf{Knowledge Prediction}}
\textbf{\#Input}: \begin{CJK}{UTF8}{gbsn}函数\$f(x)\$是定义在\$R\$上的奇函数,且对任意\$x \textbackslash in R\$都有\$f(x+6)=f(x)+f(3-x)\$,则\$f(2010)\$的值为\$\textbackslash SIFBlank\$ \end{CJK} \newline
\textbf{\#Output}: Knowledge=\begin{CJK}{UTF8}{gbsn}[三角函数, 三角函数及其恒等变换, 半角的三角函数]\end{CJK}
\\
\hline
\centerline{\textbf{Difficulty Prediction}}
\textbf{\#Input}: \begin{CJK}{UTF8}{gbsn}若连续掷两次骰子, 第一次掷得的点数为\$m\$,第二次掷得的点数为\$n\$, 则点\$P(m,n)\$落在圆\$x\^{}\{2\} + y\^{}\{2\} = 16\$ 内的概率是 \$\textbackslash SIFBlank\$	\end{CJK} \newline
\textbf{\#Output}: Difficulty=0.6316	\\
\hline
\centerline{\textbf{Discrimination Prediction}}
\textbf{\#Input}: \begin{CJK}{UTF8}{gbsn}若连续掷两次骰子, 第一次掷得的点数为\$m\$,第二次掷得的点数为\$n\$, 则点\$P(m,n)\$落在圆\$x\^{}\{2\} + y\^{}\{2\} = 16\$ 内的概率是 \$\textbackslash SIFBlank\$	\end{CJK} \newline
\textbf{\#Output}: Discrimination=0.4153 \\
\hline
\centerline{\textbf{Quality Prediction}}
\textbf{\#Input}: \begin{CJK}{UTF8}{gbsn}设\$A\$，\$B\$是椭圆\$c:  \textbackslash frac\{x\^{}2\}\{3\} + \textbackslash frac\{y\^{}2\}\{m\} = 1\$长轴的两个端点，若\$C\$上存在点\$M\$满足\$\textbackslash  angle \textbackslash mathrm\{AMB\}=120\^{}\{\textbackslash circ\}\$，则\$m\$的取值范围是\$\textbackslash SIFBlank\$\end{CJK} \newline
\textbf{\#Output}: score=5 (highest ratings), label=2 (exam problems)\\
\hline
\centerline{\textbf{Similarity Prediction}}
\textbf{\#Input}: \newline
Problem1=\begin{CJK}{UTF8}{gbsn}已知\$ \textbackslash sin \textbackslash left( \textbackslash frac\{\textbackslash pi\}\{4\} - x \textbackslash right) = \textbackslash frac\{1\}\{3\} \$，\$\textbackslash sin 2 x\$的值为\$\textbackslash SIFBlank\$ \end{CJK} 	\newline
 Problem2=\begin{CJK}{UTF8}{gbsn}已知\$ \textbackslash sin \textbackslash left( \textbackslash theta + \textbackslash frac\{\textbackslash pi\}\{2\} \textbackslash right) = \textbackslash frac\{3\}\{5\} \$，则\$\textbackslash cos 2\textbackslash theta\$等于\$\textbackslash SIFBlank\$ \end{CJK} 	\newline
\textbf{\#Output}: Similarity=0.733  \\
\hline
\end{tabular}}
\end{table}

\begin{table*}[h]
\setlength{\abovecaptionskip}{1pt}
\caption{Examples of SIF items of different subjects.}
\label{tab:app_sif_sample}
\centering
\resizebox{\linewidth}{!}{
\begin{tabular}{|c|p{14cm}|}
\hline 
\textbf{English} &
\textbf{content}: \begin{CJK}{UTF8}{gbsn} Knowing what colors look good \$\textbackslash SIFBlank\$ your skin is of great importance when you buy clothes.\end{CJK} \newline \textbf{options}: \begin{CJK}{UTF8}{gbsn}A:beyond, B:within, C:against, D:over\end{CJK} \newline \textbf{answer}: \begin{CJK}{UTF8}{gbsn}\$C\$\end{CJK} \newline \textbf{knowledge}: \begin{CJK}{UTF8}{gbsn}Vocabulary, Pronunciation, Phrases\end{CJK} \quad \textbf{difficulty}: \begin{CJK}{UTF8}{gbsn} 0.569\end{CJK}
\\
\hline
\textbf{Math} &
\textbf{content}: \begin{CJK}{UTF8}{gbsn}\$\textbackslash SIFChoice\$函数\$f(x)\$是定义在\$R\$上的奇函数,且对任意\$x \textbackslash in R\$都有\$f(x+6)=f(x)+f(3-x)\$,则\$f(2010)\$的值为\$\textbackslash SIFChoice\$ \end{CJK} \newline \textbf{options}: \begin{CJK}{UTF8}{gbsn}A:\$2010\$, B:\$-2010\$, C:\$0\$, D:不确定\end{CJK} \newline \textbf{answer}: \begin{CJK}{UTF8}{gbsn}\$C\$\end{CJK} \newline \textbf{knowledge}: \begin{CJK}{UTF8}{gbsn}三角函数, 三角函数及其恒等变换, 半角的三角函数\end{CJK} \quad \textbf{difficulty}: \begin{CJK}{UTF8}{gbsn} 0.396\end{CJK}
\\
\hline
\textbf{Chinese} &
\textbf{content}: \begin{CJK}{UTF8}{gbsn}对下面文段中加点词的解释有错误的一项是\$\textbackslash SIFChoice\$ 蒋氏大戚,汪然出涕曰:“君将哀而\$\textbackslash textf{生,u}\$之乎?则吾\$\textbackslash textf{斯,u}\$役之不幸,未若复吾赋不幸之甚也,\$\textbackslash textf{向,u}\$吾不为斯役,则久已\$\textbackslash textf{病,u}\$矣。……”\end{CJK} \newline \textbf{options}: \begin{CJK}{UTF8}{gbsn}A:君将哀而\$\textbackslash textf{生,u}\$之乎生:使……活下去, B:则吾\$\textbackslash textf{斯,u}\$役之不幸斯:此,这, C:\$\textbackslash textf{向,u}\$吾不为斯役向:从前, D:则久已\$\textbackslash textf{病,u}\$矣病:患病\end{CJK} \newline \textbf{answer}: \begin{CJK}{UTF8}{gbsn}\$D\$\end{CJK} \newline \textbf{knowledge}: \begin{CJK}{UTF8}{gbsn}None\end{CJK} \quad \textbf{difficulty}: \begin{CJK}{UTF8}{gbsn} 0.053\end{CJK}
\\
\hline
\textbf{Physics} &
\textbf{content}: \begin{CJK}{UTF8}{gbsn}如图所示,在一个凸透镜的主光轴上距光心\$2\$倍焦距的地方放一点光源\$S,\$在透镜的另一侧距光心\$3\$倍焦距的地方垂直于主光轴放置的屏上得到一个亮圆斑,若将透镜的上半部遮住,则\$\textbackslash SIFChoice\$ \$\textbackslash FigureID{001b16ae-e7af-11eb-be0e-40f2e9c851ac}\$ \end{CJK} \newline \textbf{options}: \begin{CJK}{UTF8}{gbsn}A:屏上亮圆斑上半部消失, B:屏上亮圆斑下半部消失, C:屏上亮圆斑仍完整,但亮度减弱, D:屏上亮圆斑直径减小\end{CJK} \newline \textbf{answer}: \begin{CJK}{UTF8}{gbsn}\$C\$\end{CJK} \newline \textbf{knowledge}: \begin{CJK}{UTF8}{gbsn}None\end{CJK} \quad \textbf{difficulty}: \begin{CJK}{UTF8}{gbsn} 0.732\end{CJK}
\\
\hline
\textbf{Chemistry} &
\textbf{content}: \begin{CJK}{UTF8}{gbsn}某化合物在氧气中完全燃烧后生成二氧化碳和水,已知生成的二氧化碳和水的质量比为\$22\$:\$9\$,\$则该化合物的化学式可能是\$\textbackslash SIFChoice\$\end{CJK} \newline \textbf{options}: \begin{CJK}{UTF8}{gbsn}A:\$C\_{2}H\_{2}\$, B:\$C\_{2}H\_{4}\$, C:\$C\_{2}H\_{5}OH\$, D:\$CH\_{4}\$\end{CJK} \newline \textbf{answer}: \begin{CJK}{UTF8}{gbsn}\$B\$\end{CJK} \newline \textbf{knowledge}: \begin{CJK}{UTF8}{gbsn}None\end{CJK} \quad \textbf{difficulty}: \begin{CJK}{UTF8}{gbsn} 0.496\end{CJK}
\\
\hline
\textbf{Biology} &
\textbf{content}: \begin{CJK}{UTF8}{gbsn}下列关于人体血糖平衡调节的叙述中,正确的是\$\textbackslash SIFChoice\$\end{CJK} \newline \textbf{options}: \begin{CJK}{UTF8}{gbsn}A:胰岛细胞产生的激素均能降低血糖浓度, B:胰岛\$\textbackslash alpha\$细胞和\$\textbackslash beta\$细胞分泌的激素协同调节血糖平衡, C:细胞摄取和利用血液中葡萄糖的能力下降,可导致血糖升高, D:糖尿病是由于经常摄入过量的糖所引起的,所以停止吃糖不用药就可治愈\end{CJK} \newline \textbf{answer}: \begin{CJK}{UTF8}{gbsn}\$C\$\end{CJK} \newline \textbf{knowledge}: \begin{CJK}{UTF8}{gbsn}None\end{CJK} \quad \textbf{difficulty}: \begin{CJK}{UTF8}{gbsn} 0.496\end{CJK}
\\
\hline
\textbf{History} &
\textbf{content}: \begin{CJK}{UTF8}{gbsn}1902年2月﹣4月，梁启超又在《新民丛报》上发表《论民族竞争之大势》，指出：“今日欲救中国，无他术焉，亦先建设一民族主义之国家而已。”为此，梁启超 \$\textbackslash SIFChoice\$\end{CJK} \newline \textbf{options}: \begin{CJK}{UTF8}{gbsn}A:主张国家属于人民，王侯将相是人们的公仆, B:借助经学的外衣否定君主专制, C:宣传伸民权、设议院、变法图存, D:主张民主与科学，反对专制与愚昧、迷信\end{CJK} \newline \textbf{answer}: \begin{CJK}{UTF8}{gbsn}\$C\$\end{CJK} \newline \textbf{knowledge}: \begin{CJK}{UTF8}{gbsn}近代中国, 近代中国的思想潮流, 维新思想\end{CJK} \quad \textbf{difficulty}: \begin{CJK}{UTF8}{gbsn} 0.295\end{CJK}
\\
\hline
\textbf{Geography} &
\textbf{content}: \begin{CJK}{UTF8}{gbsn}下列实例中,证明了地球上海陆发生过变迁的是\$\textbackslash SIFChoice\$ \end{CJK} \newline \textbf{options}: \begin{CJK}{UTF8}{gbsn}A:流水堆积成的三角洲, B:风力侵蚀成的蘑菇石, C:经常发生地震现象, D:在喜马拉雅山中发现了海洋生物化石\end{CJK} \newline \textbf{answer}: \begin{CJK}{UTF8}{gbsn}\$D\$\end{CJK} \newline \textbf{knowledge}: \begin{CJK}{UTF8}{gbsn}None\end{CJK} \quad \textbf{difficulty}: \begin{CJK}{UTF8}{gbsn} 0.805\end{CJK}
\\
\hline
\end{tabular}}
\vspace{-2mm}
\end{table*}



For each downstream task, we construct high-quality tagged datasets. Table \ref{tab:app_downstream_sample} provides illustrative data examples for each downstream task within the math subject domain.

Though we have introduced our downstream tasks in Section~\ref{subsec:eval}, more detailed descriptions of these tasks and our settings are as follows.
\begin{itemize}
    \item \textbf{Knowledge Prediction:} The related knowledge concepts are organized into a 3-level knowledge hierarchy in our case, where knowledge concepts in lower-layers are considered as subclasses of those in upper-layers. We adopt different pre-trained language models as encoders of questions, and feed the extracted semantic representations into HARL~\cite{huang2019hierarchical} for knowledge prediction, which is the network specifically designed for hierarchical classification.
    \item \textbf{Difficulty prediction:} The difficulty score is calculated with real-world student response data with classical test theory~\cite{devellis2006classical}. Specifically, we use 1-pass\_rate of each question as its difficulty. A higher difficulty means that the question is harder and fewer students can answer correctly. We use a simple linear layer with a Sigmoid function as the prediction network.
    \item \textbf{Discrimination prediction:} Following classical test theory, the discrimination score is formally defined as $D = P_u - P_i$, where $P_u$ is the correct response rate of the group with upper ability and $P_i$ represent the correct response rate of the lower group. It measures whether and how higher-ability students performing better than lower-ability students. We use a simple linear layer with a Sigmoid function as the prediction network. 
    \item \textbf{Quality prediction:} We annotate the quality scores in a semi-automatic way based on an assumption that real exam papers, which are designed by expert committees and contain comprehensive questions, have higher scores than mock and practice papers. We annotate real exam papers, mock papers and practice papers with basic ratings of 5, 4 and 3 respectively.
    Lower-quality samples are generated by randomly removing formulas, images, or text from question contents, which will be rated from 1 to 4 based on different portions of removal. For modeling, we employ two simple linear layers with Softmax and Sigmoid functions for scenario classification and rating regression, respectively. The metric accuracy evaluates the scenario classification and MSE evaluates the quality score prediction.
    \item \textbf{Similarity prediction:} We invited three experts to label the similarities of question pairs from 0 to 10, which will be scaled to 0-1. When determining the similarities of question pairs, experts are expected to consider the similarity of their related knowledge, required ability and quality. We directly used the pre-trained language models as frozen encoders for question representations without any further fine-tuning, and use the cosine similarity of representations as the predicted similarity.
\end{itemize}


\end{document}